# Pick the Right Edge Device: Towards Power and Performance Estimation of CUDA-based CNNs on GPGPUs

Christopher A. Metz[1]   Mehran Goli[1,2]   Rolf Drechsler[1,2]
[1]Institute of Computer Science, University of Bremen, 28359 Bremen, Germany
[2]Cyber-Physical Systems, DFKI GmbH, 28359 Bremen, Germany
{cmetz, mehran, drechsler}@uni-bremen.de

*Abstract*—The emergence of *Machine Learning* (ML) as a powerful technique has been helping nearly all fields of business to increase operational efficiency or to develop new value propositions. Besides the challenges of deploying and maintaining ML models, picking the right edge device (e.g., GPGPUs) to run these models (e.g., CNN with the massive computational process) is one of the most pressing challenges faced by organizations today. As the cost of renting (on Cloud) or purchasing an edge device is directly connected to the cost of final products or services, choosing the most efficient device is essential. However, this decision making requires deep knowledge about performance and power consumption of the ML models running on edge devices that must be identified at the early stage of ML workflow.

In this paper, we present a novel ML-based approach that provides ML engineers with the early estimation of both power consumption and performance of CUDA-based CNNs on GPGPUs. The proposed approach empowers ML engineers to pick the most efficient GPGPU for a given CNN model at the early stage of development.

*Index Terms*—power consumption, performance estimation, GPGPU, CUDA, machine learning, neural networks

## I. INTRODUCTION

Nowadays, *Machine Learning* (ML) has become very important for all types of industries, ranging from manufacturing to scientific-, health- and security-related applications [1]–[3]. This increasing deployment of ML is not only for big companies such as Google, Amazon, or Facebook but also for small and medium-sized enterprises (SME). For example, about 35% of the SME in Germany have already used ML for their applications [4].

One of the most important use cases of ML (e.g., in the German industry) is automatization and smart sensors [5]. Among the existing ML algorithm, *Convolutional Neural Networks* (CNNs) is the most popular ML algorithm for image recognition in automated manufacturing process control [6]. The convolutional layers, made up of 4-dimensional convolutions, are responsible for over 90% of the computation and require processing massive amounts of data with potentially trillions of computations per second [7]. Due to this massive computation, ML engineers take advantage of edge devices such as GPGPUs to gain performance and meet the time-to-market constraints. However, edge devices such as GPGPUs come with a wide variety of series, some of them can be very expensive and consume too much power. Therefore, for ML engineers or companies that want to run their ML algorithms such as CNNs on a GPGPU, the selection of an underlying edge device is very essential.

According to [4], statistics illustrate that 69% of the small to medium businesses face some initial issues when adopting ML for their applications. The statistics illustrate that lacking knowledge about finding the proper underlying edge device which most fit to given ML algorithms is the main challenge of such business.

For example, consider a scenario that company `AA` rents a GPGPU from company `BB` to run its CNN algorithm. To achieve the maximum performance, `AA` chooses the most expensive GPGPU of the BB company. This selection directly affects the cost of the final product of `AA` for which this CNN is used. However, there could be another alternative that `AA` could choose to gain the same performance at a lower cost.

In the case of using CNN, the fundamental concern that should overcome first is to find the most power- and performance-efficient edge device that fits it most. This is considered a critical step in using such ML models as under approximation of the proper device causes performance loss and increases the time-to-market constraints. On the other hand, over-approximation of the required edge devices is directly connected to paying more money to either purchase or rent them in a Cloud. Moreover, in the case of online applications such as smartphones, this over-approximation of GPGPU selection can reduce the battery lifetime (as the power consumption increases) and increase the area overhead.

In this paper, we focus on this scenario. We propose a novel approach to estimate the power consumption and performance of a given CUDA-based CNN model on GPGPUs. Our proposed approach statically analyzes the CNN instructions and GPGPU architectural information and take advantage of machine learning techniques for its estimation process. By this, ML engineers are able to choose the most efficient GPGPU for their CNN model at the early stage of development.

The structure of this paper is as follows: Section II presents the related work in this domain. The proposed methodology is described in Section III. The benchmarks and evaluation methods are specified in Section IV. The paper is concluded in Section V.

This work was supported in part by the German Federal Ministry of Education and Research (BMBF) within the project SATiSFy under contract no. 16KIS0821K, by the Data Science Center of the University of Bremen (DSC@UB), and by the University of Bremen's graduate school SyDe, funded by the German Excellence Initiative.

## II. RELATED WORK

The methods in [8], [9] focus on estimating the power consumption of CUDA-based applications at run-time. [8] creates a neural network that receives the CUDA instructions and the global, local, shared, and texture memories as inputs. The output of the neural network is power consumption. In contrast to [8], [9] uses more GPGPU architecture information to calculate power consumption. The power consumption estimation algorithm is based on twelve different GPGPU components (e.g. FP32-ADD/MUL/FMA, INT, SF, and CF units). However, the main limitation of these methods is their dependency on the run time information for measuring power consumption. It means that they cannot predict for a given application (e.g. CNN), the amount of power consumption without executing the model. Our approach extends the list of GPGPU components that affects performance and power consumption. Moreover, we focus on the power and performance estimation of a given CNN in order to help ML engineers to pick the most appropriate GPGPUs.

The method in [10] introduces a statistical power model based on the CUDA performance counters. The method in [11] takes advantage of machine learning techniques and the CUDA performance counters to predict the performance and power consumption of GPGPUs across a range of hardware configurations. However, using the CUDA performance counters limits the model to only predict based on run-time data. For example, while [11] can be used to find the best configuration for a specific device, it cannot be used to choose the most efficient device in the early design stage before running the application.

The method in [12] uses tree-based regression to predict the power consumption of CUDA based applications on GPGPUs. It analyzes the GPGPU architecture and measures the power consumption of the PTX instructions. As the method takes advantage of GPGPUSim for the simulation and modeling of GPUGP power consumption, the amount of PTX instructions is limited. Moreover, the power measurements are not of real hardware as they use a simulator instead of real GPGPUs.

The method in [13] considers scaling frequencies for core and memory frequency of GPGPUs. In [14], a machine learning model is introduced that estimates the performance of CPU code before porting to GPGPU code, so it is possible to decide if executing on GPGPU gives a performance boost before writing the GPGPU code. In [15], a similar goal is pursued but the speedup prediction of the code can be performed on GPGPU.

Our approach focuses on machine learning-based estimation in the early design stage before running the application (i.e., a given CNN) on any target edge systems.

## III. POWER AND PERFORMANCE PREDICTION METHODOLOGY

In order to predict the power consumption and the performance of a given CNN before executing it on the target GPGPU, it is important to focus on features that are known at the early stages of system design. These features are related to the GPGPU architectural components that affect performance and power consumption. In contrast to [8], [9] we focus on the hardware components and not on performance counter or events. Similar to [8], we also consider the instructions that are executed by the GPGPU. By this, it is possible to estimate the power consumption of executing CNNs on different GPGPUs.

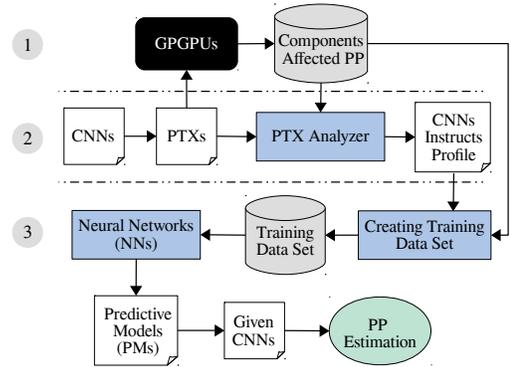

Fig. 1. Overview of the proposed methodology.

### A. Methodology Overview

The proposed methodology is divided into three main steps shown in Fig. 1. In the first step, we perform an analysis to extract those GPGPU components that have an impact on the overall power consumption and performance of the GPGPU if they are loaded by the running (e.g. CNN) application instructions.

In the second step, we translate CNN models into a set of PTX instructions and analyze the generated PTX file to extract the instructions that are loaded in GPGPU components to run the CNN. By this, we only consider those GPGPU components that directly affect the overall performance and power consumption when CNN is run.

In the third step, we build a training dataset based on the CNN instructions, the amount of power consumption for each instruction and the GPGPU components that are loaded by the CNN instructions. In the case of performance estimation, the training data set includes the number of CNN instructions, GPGPU components that are loaded by the CNN instructions, and the number of instructions that can be rendered by the GPGPU per second. Once the training dataset is created, we take advantage of ML algorithms (i.e. neural networks) to train the data set and create a predictive model. The CNN instructions and the GPGPU components that are loaded by them are the inputs while the overall performance and power consumption of the GPGPU are the outputs. Finally, the generated predictive model is used to estimate the performance and power consumption of a given CNN at the development process when GPGPU is the target device.

### B. GPGPUs Architecture Analysis

The performance and power consumption of the GPGPUs are affected by many factors. Nvidia lists all the components of the GPGPU's architecture in their white paper [16], [17]. We focus on components that are available over the various Nvidia architectures, e.g. amount of streaming multiprocessors, clock frequency for compute units and memory, L2 Cache size. This enables our approach to estimate the power consumption

```
1   ld.param.u64    %rd1, [copy_5_param_0];
2   ld.param.u64    %rd2, [copy_5_param_1];
3   cvta.to.global.u64  %rd3, %rd2;
4   cvta.to.global.u64  %rd4, %rd1;
5   mov.u32         %r1, %ctaid.x;
6   mov.u32         %r2, %tid.x;
7   shl.b32         %r3, %r1, 8;
8   or.b32          %r4, %r3, %r2;
9   shl.b32         %r5, %r4, 2;
10  or.b32          %r6, %r5, 1;
11  or.b32          %r7, %r5, 2;
12  or.b32          %r8, %r5, 3;
13  cvt.u16.u32     %rs1, %r4;
14  mul.wide.u16    %r9, %rs1, -7281;
15  shr.u32         %r10, %r9, 22;
16  mul.wide.u32    %rd5, %r8, 954437177;
17  shr.u64         %rd6, %rd5, 33;
18  cvt.u32.u64     %r11, %rd6;
19  and.b32         %r12, %r11, 31;
20  mul.wide.u32    %rd7, %r8, -1431655765;
```

Fig. 2. Part of a PTX file of a CNN model.

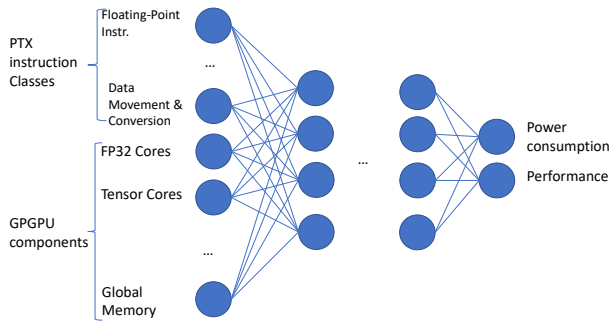

Fig. 3. Neural network architecture for power and performance prediction.

through different GPGPU models. For example, we do not consider the FP64 Cores of the Nvidia V100 [16] as there is no information in [17] for other GPGPU architecture related to these components. Moreover, to know those GPGPUs' components that directly affect the *Performance and Power consumption* (PP) of GPGPUs, we run the compiled PTX files of various CNN algorithms on different GPGPUs. Then for each PTX instruction type (which is loaded to a specific GPGPU component), we measured the share of those components in comparison to the total performance and power consumption of the GPGPU. By this, GPGPU components that directly are related to the performance and power factors are extracted, and results can even be reduced more to cover only the most important components.

### C. PTX Instructions Analysis

Nvidia offers the CUDA Library to run applications on their GPGPUs. However, frameworks like Tensorflow already including the CUDA Library so that the user can easily run their application on GPGPUs [18]. To run the CUDA code on GPGPU, the code is compiled to the *Parallel Thread Execution* (PTX) which contains the machine language instructions. These low-level instructions include every memory access (read and write) as well as computational instructions such as ADD, MUL, FMA and etc. [19].

This detailed low-level information enables us to perform an analysis on the generated PTX files (obtained from the CNNs compilation results) to categorize the instructions and create several unique classes based on how they load GPGPUs components. Some classes are: *Comparisons and selection instructions*, *Floating-Point instructions*, *Arithmetic instructions* and more. We use this classification of PTX instructions and count the number of instructions for each class. We only consider these types which are needed over different CNNs.

For example, Fig. 2 shows part of a simple PTX file. This PTX file includes eight *data movement and conversion* (i.e., *ld*, *cvta*, *mov*, and *cvt*), three *floating-point* instructions (i.e., *mul*), nine *logic and shift* instructions (i.e., *and*, *or*, *shl*, and *shr*). Therefore, for this example, the *CNNs Instructs Profile* (Fig. 1, step 2) includes three instruction classes where for each class the number of instructions is included.

As illustrated in Fig. 1, step 2, this analysis is performed for various PTX files from different CNN algorithms. The results of this analysis are used to create a training dataset and predictive model in the next step.

### D. Creating Training Dataset and Predictive Model

In order to create a predictive model for the power and performance estimation, it is necessary to have a robust training dataset. To do this, the generated *CNNs instructs Profile* and *Components Affected PP* (Fig. 1, step 1 and 2) are used to create a training dataset w.r.t performance and power consumption. The performance of the GPGPU is measured based on the number of PTX instructions that can be run per second. The power consumption is measured in Watt with the Nvidia-smi Tool [20].

Table I shows the overall structure of the training dataset. Column *CNN and GPGPU* shows the name of the CNN algorithm and the GPGPU model (please see Section IV for detailed description). The *Classes* column lists the name of instruction classes and for each class the number of each instruction. Column *GPGPU Components* indicates the relevant GPGPUs' architectural information. The *Output* column presents the total performance and power consumption for each benchmark (CNNs and GPGPUs combination). Thus, each row of the table indicates an observation in the training dataset where the first three columns are the inputs (or features) while the total power consumption and performance (column *Output*) are the outputs. The training dataset is split into 70% for the training phase and 30% for the validation phase.

To predict total power consumption and performance (i.e., Watt and instruction per cycle) of a given CNN, a set of *Neural Networks* (NNs) are designed with different layers. NNs take as input the features shown in table I and predict the outputs (power and performance). Fig. 3 illustrates an abstract view of the neural network. The exact amount of neurons in each layer depends on the number of PTX instruction classes and GPGPU components. To evaluate the predictive models, we consider the CNNs and GPGPUs which are not in the training dataset.

TABLE I
EXAMPLE OF TRAINING DATASET STRUCTURE USED TO CREATE THE PREDICTIVE MODELS

| CNN and GPGPU | Classes | | | | GPGPU Components | | | | Output | |
|---|---|---|---|---|---|---|---|---|---|---|
| | Data movement and conversion | Floating-Point | ... | Class N | L2 Cache | FP32 Cores | ... | SM | Power Consumption | Performance |
| CNN (Fig. 2) + V100 | 8 | 3 | ... | ... | 6144KB | 5120 | ... | 80 | $Power_1$ | $Performance_1$ |
| CNN (Fig. 2) + 2080Ti | 8 | 3 | ... | ... | 5632KB | 4352 | ... | 68 | $Power_2$ | $Performance_2$ |
| CNN (Fig. 2) + 1080Ti | 8 | 3 | ... | ... | 2816KB | 3584 | ... | 28 | $Power_3$ | $Performance_3$ |
| $CNN_n$ + $GPGPU_m$ | ... | ... | ... | ... | ... | ... | ... | ... | $Power_{n*m}$ | $Performance_{n*m}$ |

## IV. BENCHMARKS AND METHODOLOGY EVALUATION

In order to evaluate the proposed methodology, a set of benchmarks is considered including three different types of GPGPUs as well as different pre-trained well-known CNN algorithms. The GPGPUs that we considered in our benchmark are based on different architectures. This enables us to develop the proposed methodology more generic in the prediction of performance and power consumption of running applications on GPGPUs. The evaluation of the proposed methodology is performed by estimating the performance and power consumption of a given CNN (which is not in the training dataset) using the predictive models. The CNN is run on the GPGPU while the *Nvidia-smi Tool* is used to measure the power consumption of the GPGPU. Nvidia-smi reports the power consumption in a *csv* file. For this, a measurement interval of one second is considered (i.e., every second a measurement is written to the csv file). The power consumption is averaged over the entire run and compared to the ML model estimation.

## V. CONCLUSION AND FUTURE WORKS

We propose a novel ML-based approach to predict power consumption and performance of CUDA-based CNNs on GPGPUs. This enables ML engineers to pick the GPGPU that suits best their CNN in terms of performance and power consumption. The results of this analysis help ML engineers create their systems more efficiently and avoid high productions cost.

The proposed approach is based on analyzing the PTX instructions obtain from various standard CNN models and the GPGPUs' architectural information. We take advantage of this information to build a set of training data and use them to generate a predictive model to estimate the performance and power consumption of a given CNN. The proposed methodology is under development and is in the early evaluation phase. We believe this new line of research can truly help designers to make better decisions in the early stage of selecting edge devices. Moreover, to help them in optimizing their ML algorithm to fulfill the constraints of a given edge device.

As future works, we plan to consider more benchmarks (i.e., GPGPUs series and CNN algorithms) to make our predictive models more accurate. The other direction is to consider not only the CNN algorithms but also expand our method for other ML methods such as *Neural Networks* (NNs). Moreover, it is also interesting to consider other edge devices such as *Tensor Processing Units* (TPUs) to provides ML engineers with more choosing options to run their ML algorithms.